\documentclass{article}
\usepackage{spconf,amsmath,graphicx}
\usepackage[normalem]{ulem}

\title{Localizing firearm carriers by identifying human-object pairs}
\name{Abdul Basit, Muhammad Akhtar Munir, Mohsen Ali, Arif Mahmood}
\address{Department of Computer Science,\\
	Information Technology University (ITU), 346-B, Ferozepur Road, Lahore, Pakistan.\\
	Emails: \{abdul.basit, akhtar.munir, mohsen.ali, arif.mahmood\}@itu.edu.pk }
  
\begin{document}
%
\maketitle
\begin{abstract}
Visual identification of  gunmen in a crowd is a challenging problem, that requires resolving the association of a person with an object (firearm). 
We present a novel approach to address this problem, by defining human-object interaction (and non-interaction) bounding boxes. 
In a given image, human and firearms are separately detected. 
Each detected human is paired with each detected firearm, allowing us to create a \textit{paired} bounding box that contains both object and the human.
A network is trained to classify these paired-bounding-boxes into human carrying the identified firearm or not. 
Extensive experiments were performed to evaluate effectiveness of the algorithm, including exploiting full pose of the human, hand-keypoints, and their association with the firearm. 
The knowledge of spatially localized features is key to success of our method by using multi-size proposals with adaptive average pooling. 
We have also extended a previously firearm detection dataset, by adding more images and  tagging in extended dataset the human-firearm pairs (including bounding boxes for firearms and gunmen). 
The experimental results (\textit{78.5 $AP_{hold}$}) demonstrate  effectiveness of the proposed  method.

\end{abstract}
\section{Introduction}
\label{sec:intro}

Gun violence incidents are universal across the globe. These incidents account for many lives around the world every year \cite{newyear, Americas23:bbc, amnesty}. A lot of administrative steps have been taken to minimize these incidents but despite all the efforts, the frequency of such events is increasing with time. Due to lack of direction in efficient technology based systems for firearm control on large scale, number of casualties have grown over the years. There is an immense need to come up with a scientific solution to this problem, specially with the huge number of cameras and other imaging systems available today.

Many countries and private security agencies have deployed surveillance systems in public places like schools, colleges, parks, and malls for public safety. These systems require huge workforce to manually look at the footage from the surveillance feed. The downside of operating manually is that it requires  lots of human efforts with more chances of mistakes, because of lack of vigilance over long working hours. So, there is a need for a method that should not only be able to detect firearms but it should also be able to categorize the human, who is carrying. Such a novel scientific solution can be embedded in surveillance systems for significant improvement in identifying a potential gun violence incident.

\begin{figure}[t]
\begin{minipage}[b]{1.0\linewidth}
  \centering
  \centerline{\includegraphics[width=8.5cm]{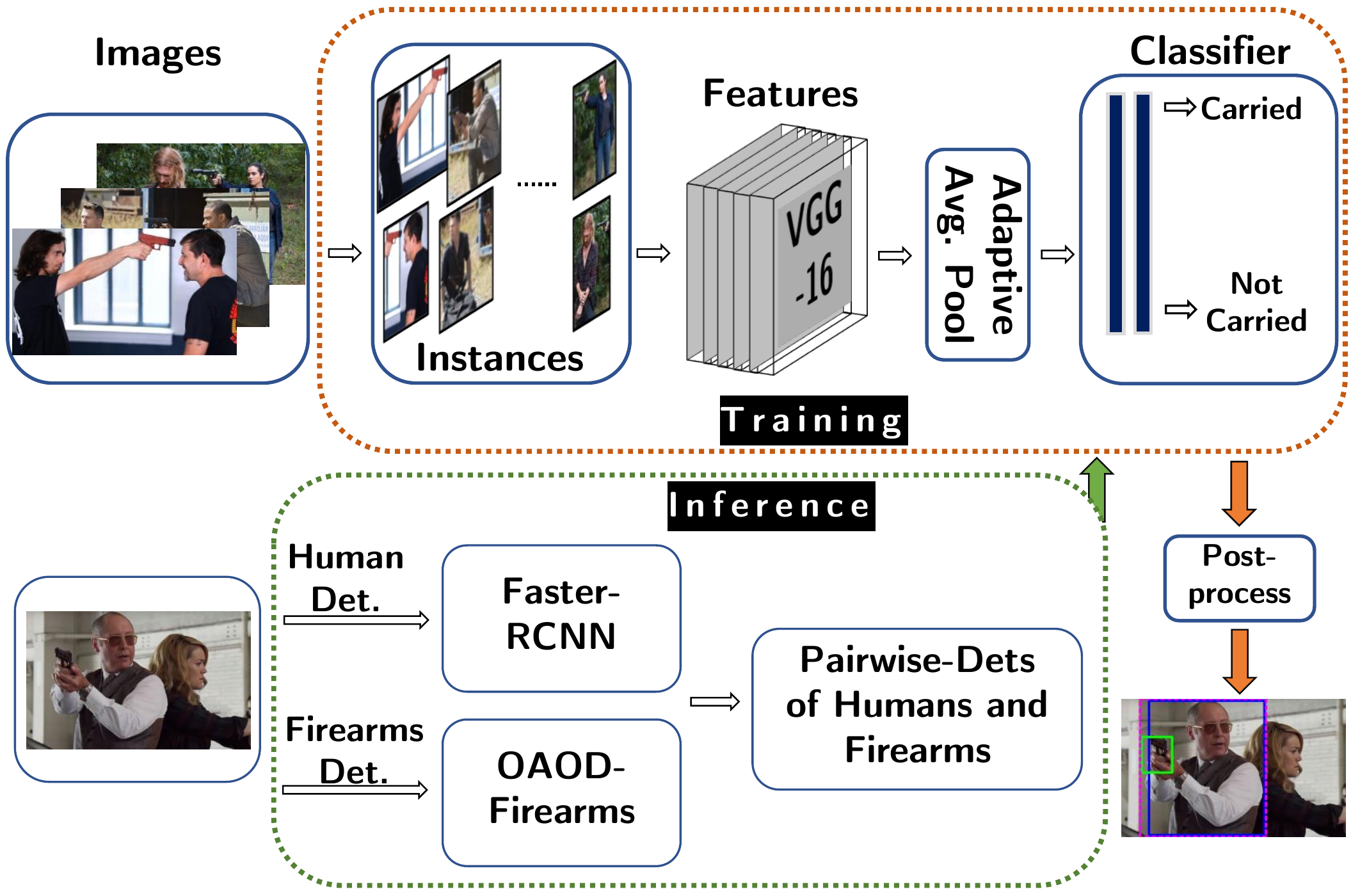}}
\end{minipage}
\caption{Proposed firearm carrier detection network which not only tells the firearm is carried but also identifies the carrier at image level. Use of Adaptive Average Pooling allow us to train our model using multi-size proposals. }
\label{fig:main}
\vspace{-5mm}
\end{figure}

In the recent years, use of visual systems for classification \cite{huang2017densely,he2016deep,Simonyan15} and detection \cite{ren2015faster,redmon2018yolov3,redmon2017yolo9000,redmon2016you,liu2016ssd,lin2017feature,fu2017dssd,tian2019fcos,duan2019centernet} are of great importance. Deep learning based neural networks \cite{huang2017densely,he2016deep,Simonyan15} have performed quite well in key computer vision tasks.
These systems learn unique visual features associated with different objects from the given set of images.
Many object detection algorithms are designed to perform relatively well on benchmark datasets like PASCAL VOC \cite{everingham2010pascal} and MS-COCO \cite{lin2014microsoft}, but to apply these algorithms directly on objects like firearms is not much effective. Taking base architecture Faster RCNN \cite{ren2015faster} for object detection,  firearms (gun and rifle) detection has been improved in OAOD \cite{oaod}.

Identification of firearms is difficult task owing to the wide variety of firearms available around the globe. 
Inherent differences in shape and size make identification of firearms even more challenging. 
Furthermore, to detect if a human is holding a firearm, is a tougher task because of the clutter, small sized guns, and extremely large but thin rifles making their bounding boxes spread over a larger window which may contain multiple humans. Due to the overlapping of multiple humans in the firearm box, it is very hard to determine which human is actually holding the firearm.

Existing work still lacks to identify that who is carrying firearm? So integrating this module with firearm detection is a novel task  addressed in this work. For that we borrow subset of firearms dataset publicly available \cite{oaod} with additional images. New dataset is manually tagged containing humans and firearms (labels: gun and rifle) bounding boxes with carried and non-carried labels. Along with the dataset, in this paper, a method is proposed which detects whether the firearm is being carried by a specific human. Figure \ref{fig:main} gives a framework of final model that classifies human-firearm pair localization.

We integrate knowledge from different fields to develop a novel solution that can detect objects like firearms from OAOD and classify associations with human in images labeling carried.
Incorporating OpenPose \cite{cao2018openpose}, we use pose estimation to locate hands and build association of firearm being carried. Then as baseline results, we use OAOD for firearms detection and Faster RCNN \cite{ren2015faster} (trained on MS-COCO \cite{lin2014microsoft} with ResNet-101 \cite{he2016deep}) for human detection. Afterwards, enclosed overlap ($IoUs \geq 0.5$) between detected humans and objects (firearms) is computed and considered true positive if carried (a metric considered in Human Object Interaction (HOI) systems). To improve from baseline results, we train classification model from the association annotated in our dataset of humans and firearms being carried and not-carried. Our model is based on VGG-16 along with adaptive average pooling (AAP) to handle multi-size proposals. Moreover, AAP also allow us to capture context with maximum participation of primary object in an image.

\section{Related Work}
\label{sec:format}

\textbf{Object Detection:} With advancement of deep learning,  object detection research has achieved significant improvement  \cite{ren2015faster,redmon2018yolov3,liu2016ssd,lin2017feature, tian2019fcos, duan2019centernet}.
Generic object detection may be divided into two main categories: one stage object detectors (excel in speed) \cite{redmon2018yolov3,redmon2017yolo9000,redmon2016you,liu2016ssd, fu2017dssd, tian2019fcos,duan2019centernet} and two stage object detectors (excel in accuracy) \cite{ren2015faster, lin2017feature, singh2018analysis}. There is speed-accuracy trade-off \cite{huang2017speed}, suggesting to gain accuracy and speed at the same time is still a challenging task. 


\textbf{Firearm Detection:} 
There are much fewer visual systems  specifically dedicated to firearms detection. Javed et al. \cite{oaod} has developed a weakly supervised Orientation Aware Object (firearm) Detection system (OAOD), which detects guns and rifles in an image, based on the orientation information. Instead of taking oriented Bounding Boxes (BB) during training, axis aligned BB are used. Olmos et al. \cite{olmos2018automatic} has applied  Faster RCNN on gun detection problem and their system detects only handguns. Akcay et al. has adopted different methods including one and two stage methods for the detection of gun using x-ray baggage security imagery \cite{akcay2018using}. None of these methods presents a solution to identify firearms carriers.

\textbf{Pose Estimation:} Numerous methods have been proposed to estimate human pose in 2D and 3D vision systems \cite{simon2017hand, cao2018openpose, wei2016cpm}, but there is no work done that uses human pose estimation to predict whether a person is firearm carrier. 
OpenPose \cite{cao2018openpose} uses part affinities between joints and keypoints to estimate poses. Simon et al. proposed hands keypoint estimation using multi-view bootstrapping \cite{simon2017hand}, which estimates keypoints of hands.

\textbf{Human Object Interaction:} The work in Human Object Interaction (HOI) mainly exploits context and attention mechanism to detect human and object interaction \cite{wang2019deep}. This would not help in more complex tasks required for localizing the actor and interaction recognition.
Exploiting different human based approaches to detect human and object such as pose and action information may improve HOI performance \cite{gkioxari2018detecting}.
By integrating knowledge from different fields, we propose a novel solution that can detect objects like firearms (gun and rifle) in images along with being carried by respective human.
This paper introduces first such method which uses human and hand detection followed by firearm detection to identify firearm carriers in crowded scenes. 

\section{Baseline Carrier Detection Approaches}

In firearm detection research, no work has yet been proposed for the carrier detection. The OAOD method only detects firearms in the images, while the current work we  integrate firearm location and human information to classify firearms as being carried or not carried.  For firearm classification, consider three different baseline approaches including two based on identifying the human hand and the firearm closeness, and the third one is based on overlap of human bounding boxes and firearm bounding boxes. 

\subsection{Detecting Hands Inside Firearm Bounding-boxes (HiFB)}
\label{sect:HIFB}

In this approach firearm bounding box is detected using OAOD algorithm. 
Firearm bounding boxes are then input to Multiview Bootstrapping algorithm \cite{simon2017hand}
to identify the hand keypoints. 
From the predicted keypoints, we identify the set of the keypoints having confidence larger than a selected threshold $\alpha$. If carnality of the set is greater than the two we classify the bounding box as carried. 
In our case, $\alpha$ is set to $0.3$. 
This approach achieves reasonable accuracy for classifying gun as being carried/not-carried, however for the case of rifles, accuracy is low. It is because extremely large bounding boxes of thin and long rifles are not usually hand-centered. These large boxes cannot be used as the probable hand locations as actual hands will cover less pixels. In gun bounding boxes, this condition significantly meets the requirement and hand to box pixels ratio greater as compared to bounding boxes containing rifles. 

\subsection{Body-pose Conditioned Firearm carried detection (BCFD)}
\label{sect:BCFD}
In this approach full human body pose containing body parts is detected using OpenPose \cite{cao2018openpose} and OAOD \cite{oaod} is used to independently detect firearms in a given image.  The detected firearms bounding boxes are inspected against the estimated hand keypoints. The firearm is categorized as carried if a hand keypoints  and firearm location overlap exceeds the threshold  $\beta$.  The problem with this method is that, full human body is often not visible in most cases due to occlusions or partial appearance.  If keypoints of elbow and wrist are not detected then keypoints of hands are also remain undetected. Therefore, this approach suffers performance degradation correlated with the performance of the pose detector. 


\subsection{Measuring Overlap of Human and Firearms Bounding-boxes (OHFB)}
\label{sect:OHFB}
\begin{figure}[t]
\begin{minipage}[b]{1.0\linewidth}
  \centering
  \centerline{\includegraphics[width=8.5cm]{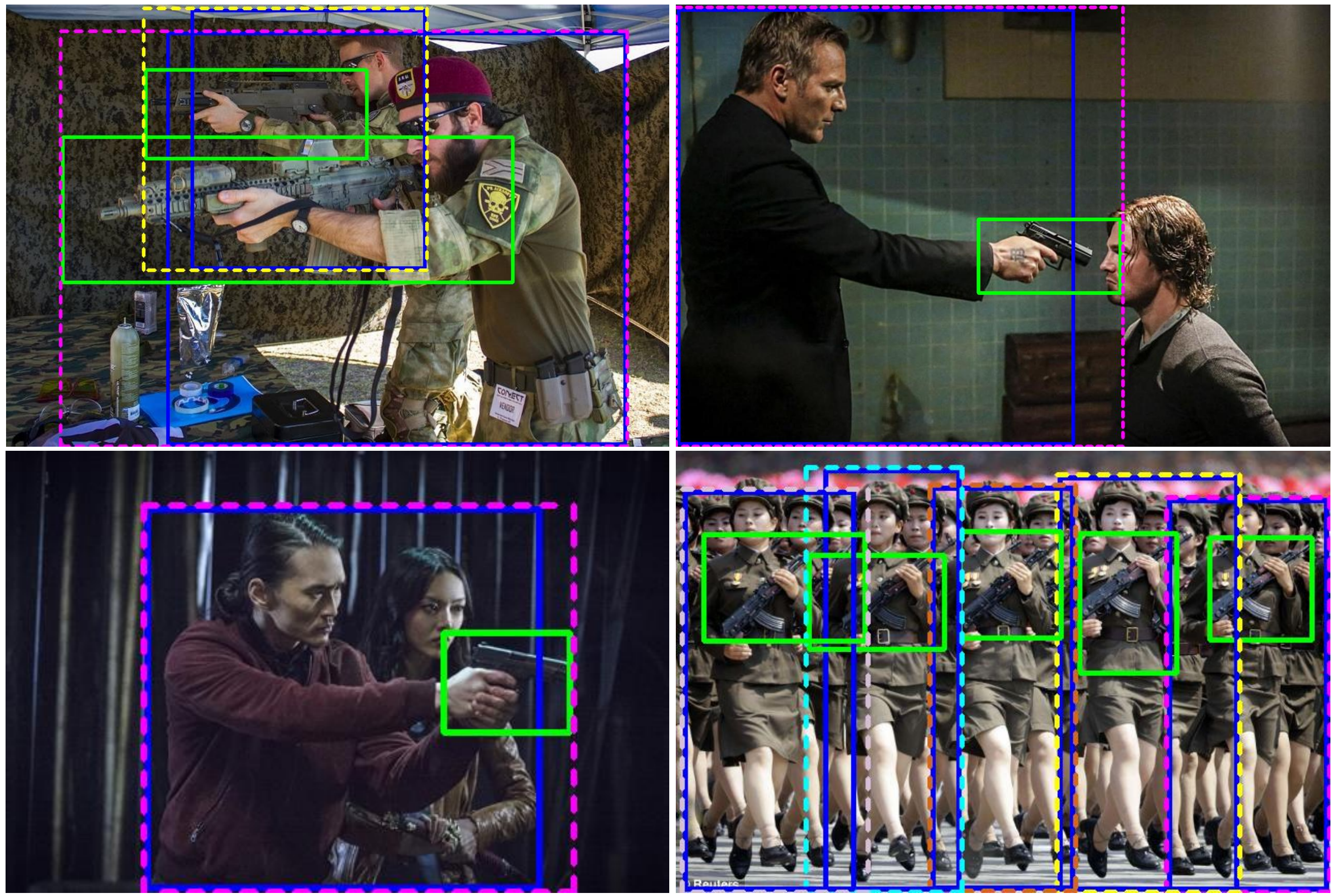}}
\end{minipage}
\caption{Output results by our proposed model. Firearms are shown in Green bounding box, humans are shown in Blue bounding box. Pairwise-extended boxes are ON with respective colors if returned carried by our classification model.}
\label{fig:results}
\vspace{-5mm}
\end{figure}

In this approach, human detection is performed by using Faster-RCNN with ResNet-101,  pre-trained on MS-COCO. The firearm detection is performed using OAOD algorithm.
Intersection of all detected firearms and detected human bounding boxes is performed.
Association between firearm and human is established by choosing one with maximum overlap.
However, associations with $IoU < 0.5$ are removed.  
This approach's performance suffers in case of crowded scene where firearm BB may have larger overlap with a non-carrier (Fig. \ref{fig:results}).

\section{Proposed Human Firearm Pair Detection (HFPD)}
\label{sec:pagestyle}


In addition to the firearm classification as carried or not carried, we aim to identify the carrier of the firearm. 
The methods (Sec.~\ref{sect:HIFB},~\ref{sect:BCFD}) where we explicitly try to create association between human pose and firearm, the pairing fails due to error in the pose estimation. 
A naive idea, considered in \ref{sect:OHFB}), is to classify firearm and human as paired on the basses of overlap between them.
Since, this method does not take in consideration the association of human body with the firearm it fails to achieve good performance, e.g. in crowded scenes.  
Therefore, we propose to train a neural-network which will learn using training data the necessary features to identify if a particular firearm and human are paired.  


In a given image, human and firearms bounding boxes are separately detected as discussed in the Section  \ref{sect:BCFD}. 
Each detected human bounding box is paired with each detected firearm, allowing us to create a \textit{paired} bounding box that contains both the firearm and the human. 
Some of these paired bounding boxes contain both firearm and its carrier, in others the pair is not associated. 
A network is trained to classify these paired-bounding-boxes into interaction or non-interaction. 
For training, the extended boxes  are manually labeled  which tell whether the human present in the box is carrying the firearm or not. Features from these multi-size pairwise-extended boxes are used to learn associations of humans and firearms. 
A classification model is trained on these instances. 
For the pair classification network, We use VGG-16 (consisting of five VGG blocks of convolutional layers) as feature extraction, followed by Adaptive Average Pooling (AAP) and two fully connected layers.
AAP allows us to use multi-size proposals without resizing them. 
The cross entropy loss function for object classification is defined as:
\begingroup
\small
\begin{align}
    L_c(p_c, g_c) = \sum_{i=1}^{n_{s}}\sum_{j=1}^{n_c} g_c(i,j) \text{log}(p_c(i,j))
\label{eq4}
\end{align}
\endgroup
where $p_c \in \mathcal{R}^{n_c}$ is the predicted class probability of being carried and not-carried and $g_c=\{\{1,0\},\{0,1\}\} \in \mathcal{R}^{n_c}$ is the ground truth class label, $n_c=2$ is the number of classes, and $n_{s}$ is the number of samples in batch.

\begin{table}[t]
\centering
\small
\caption{\small{Classification of firearm as carried or not carried}}
 \begin{tabular}{c|ccc} 
 \hline
 Methods &  Gun & Rifle & Overall \\ [0.5ex] 
 \hline
 HiFD  & 71.9 & 37.5 & 49.2 \\ \hline
 BCFD  & 51.3 & 76.4 & 66.6 \\ \hline
 \end{tabular}
 \label{tab:cls}
\end{table}



\section{Experiments and Results}
\label{sec:typestyle}

\subsection{Dataset and Annotations}
\begin{figure}[t]
\begin{minipage}[b]{1.0\linewidth}
  \centering
  \centerline{\includegraphics[width=8.5cm]{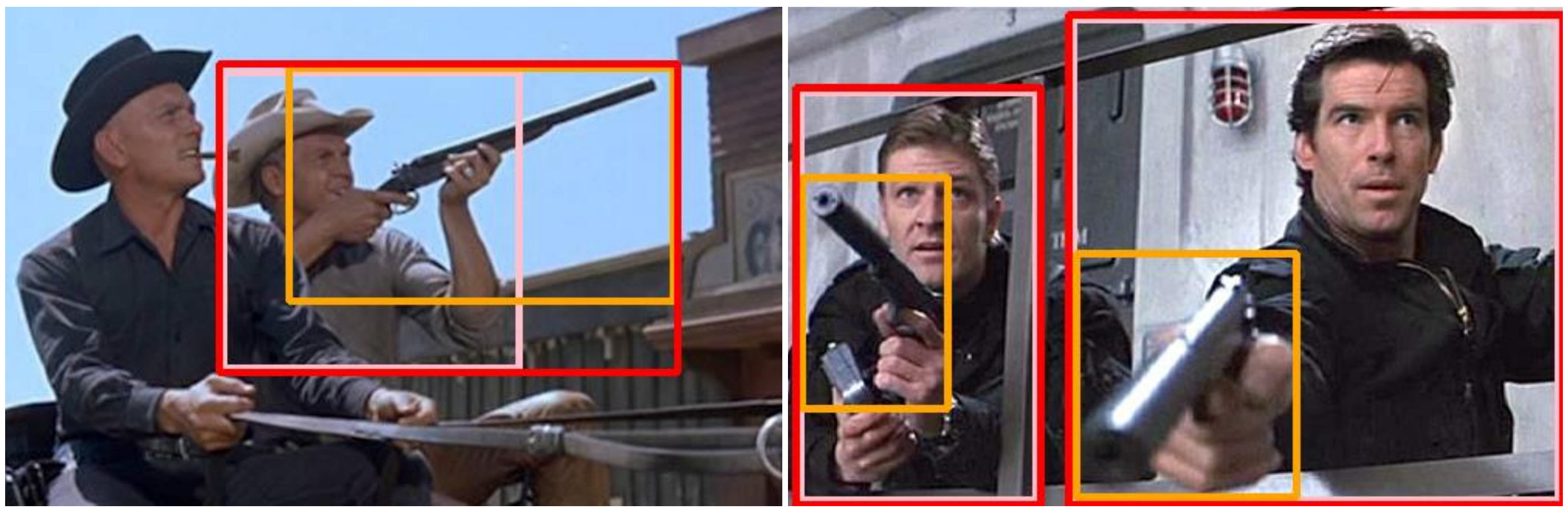}}
\end{minipage}
\caption{Samples images from the dataset, Firearms are shown in Orange bounding box, humans in Pink bounding box, Red bounding box show association of human and firearm. Negative associations are not shown to avoid clutter.}
\label{fig:dataset}
\end{figure}

As per our knowledge, we are the first to release dataset that contains bounding boxes of firearm and humans, with the association between the firearm and carrier tagged. 
A subset of images was selected from \cite{oaod}, that contain multiple humans and one or more firearms. 
900 more images were added to the dataset, bringing the size to 3128. 
For each image, firearm and human bounding box are manually annotated, including the pair-bounding box. The pair-bounding box is labelled 1 if it contains valid interaction, 0 if not.  
All the experiments are evaluated on this dataset. Some samples of images from dataset are shown in Figure \ref{fig:dataset}.

\subsection{Comparison of Different Algorithms}

We compare the performance of experiments along with the experimental details. Table \ref{tab:hfi} shows the experimental results and comparison with the baseline.
\subsubsection{Measuring Overlap of Human-Firearms Detection}
Experiments were conducted using pre-trained models, for humans (pre-trained Faster RCNN with ResNet-101 and VGG-16 on MS-COCO) and firearms (OAOD) detection. 
We share the results on both backbones, which indicate ResNet-101 achieve better detection of humans. 
The model is evaluated, end-to-end.  Both, the predicted human and firearms bounding boxes must have IoUs $\geq$ 0.5 to be considered as positive sample for the paired-bounding box.   
These results are used as baseline, for human-firearm interaction.

  

\begin{table}[t]
\centering
\footnotesize
\caption{\small{Results of Human-Firearms pair identification with and without Adaptive Average Pooling (AAP) )}}
\begin{tabular}{c|c|ccc}
\hline
\textbf{Methods}                                                  & \begin{tabular}[c]{@{}c@{}}\textbf{Backbone}\\ \textbf{HumanDet}\end{tabular} & \textbf{$AP_{Ghold}$} & \textbf{$AP_{Rhold}$} & \textbf{$AP_{hold}$} \\ \hline
\textbf{Baseline OHFB}                                                 & VGG-16            & 42.4                   & 62.6                   & 54.64                \\ \hline
\textbf{\begin{tabular}[c]{@{}c@{}}HFPD w/o\\ AAP\end{tabular}}   & VGG-16            & 62.4                   & 67.2                   & 64.7                 \\ \hline
\textbf{\begin{tabular}[c]{@{}c@{}}HFPD with\\ AAP\end{tabular}}  & VGG-16            & 64.8                   & 73.4                   & 69.3                 \\ \hline \hline
\textbf{Baseline OHFB}                                                 & ResNet-101        & 65.0                   & 74.5                   & 70.6                 \\ \hline
\textbf{\begin{tabular}[c]{@{}c@{}}HFPD w/o\\ AAP\end{tabular}}   & ResNet-101        & 72.3                   & 78.9                   & 76.0                 \\ \hline
\textbf{\begin{tabular}[c]{@{}c@{}}HFPD with \\ AAP\end{tabular}} & ResNet-101        & \textbf{75.3}                   & \textbf{81.1}                   & \textbf{78.5}                 \\ \hline
\end{tabular}
\label{tab:hfi}
\end{table}

\subsubsection{Joint Human-firearm Interaction Bounding Boxes}
For this experiment, VGG-16 network is fine-tuned (using ImageNet weights) on the annotated bounding boxes of humans-firearms association from the proposed dataset. 
Note that, we use two fully connected layers of pre-trained VGG network and add final classification layer over it. That restricts the size of image that could be input to the network. 
Adaptive average pooling (AAP) is used between last convolutional layer and first FC layer to handle multi-size inputs to the network. 
Without (w/o) AAP, results are reported with fixed resized image (224x224) that simple VGG-16 model allows but aspect ratio disturbs in most of the cases. AAP (output=7x7) is being used to handle multi-size boxes during training and testing as actual VGG-16 does not allow random sized input. AAP also helps to capture prominent details with primary object participation. It is to note that, resizing of multi-size pair-wise extended boxes are done in a way that long dimension corresponds to 600 and shorter side adjusts accordingly. It can be seen in Table \ref{tab:hfi}, that using ResNet-101 for human detection yields better proposal with firearms for our model to classify. With AAP, learning rate is 0.00001 with batch size = 1 and dropout (50\%) is used while training. We adopt stochastic gradient descent (SGD) to train the model for 20 epochs, and a momentum of 0.9.




\section{Conclusion and Future Directions}
\label{sec:majhead}

In this paper we present a novel method for localizing firearm and its carrier in a crowded scene.
The problem is not only challenging but its solutions is dearly needed in current law-and-order situation where intelligent cameras are required to perform the surveillance. 
We exploit human and firearm detection to create paired-bounding boxes for every possible human-object (firearm) pair in the image.
Reducing our problem to classifying the pair as valid interaction or not.
Employing, adaptive average pooling and multi-size proposals, we were able to achieve 78.5 $AP_{hold}$ on the test dataset. 
Extensive comparative experiments were performed by designing various baseline strategies, including ones exploring human pose and hand-keypoint information. 
In future, we explore the importance of salient areas of firearms and humans spatially correlated with their respective regions. 
Existing OAOD dataset \cite{oaod} was extended to include more challenging images and were hand-tagged to identify human-firearm pairs. 


\small
\bibliographystyle{IEEEbib}
\bibliography{refs}

\begin{thebibliography}{10}

\bibitem{newyear}
Nytimes.com,
\newblock ``In a deadly start to 2020, new-year shootings,''
  \url{https://www.nytimes.com/2020/01/01/us/new-year-shootings.html}, 2020.

\bibitem{Americas23:bbc}
BBC News,
\newblock ``America's gun culture in charts,''
  \url{https://www.bbc.com/news/world-us-canada-41488081}, 2020.

\bibitem{amnesty}
Amnesty.org,
\newblock ``Key facts about gun violence worldwide,''
  \url{https://www.amnesty.org/en/what-we-do/arms-control/gun-violence/}, 2020.

\bibitem{huang2017densely}
Gao Huang, Zhuang Liu, Laurens Van Der~Maaten, and Kilian~Q Weinberger,
\newblock ``Densely connected convolutional networks,''
\newblock in {\em Proceedings of the IEEE conference on computer vision and
  pattern recognition}, 2017, pp. 4700--4708.

\bibitem{he2016deep}
Kaiming He, Xiangyu Zhang, Shaoqing Ren, and Jian Sun,
\newblock ``Deep residual learning for image recognition,''
\newblock in {\em Proceedings of the IEEE conference on computer vision and
  pattern recognition}, 2016, pp. 770--778.

\bibitem{Simonyan15}
Karen Simonyan and Andrew Zisserman,
\newblock ``Very deep convolutional networks for large-scale image
  recognition,''
\newblock in {\em International Conference on Learning Representations}, 2015.

\bibitem{ren2015faster}
Shaoqing Ren, Kaiming He, Ross Girshick, and Jian Sun,
\newblock ``Faster r-cnn: Towards real-time object detection with region
  proposal networks,''
\newblock in {\em Advances in neural information processing systems}, 2015, pp.
  91--99.

\bibitem{redmon2018yolov3}
Joseph Redmon and Ali Farhadi,
\newblock ``Yolov3: An incremental improvement,''
\newblock {\em arXiv preprint arXiv:1804.02767}, 2018.

\bibitem{redmon2017yolo9000}
Joseph Redmon and Ali Farhadi,
\newblock ``Yolo9000: better, faster, stronger,''
\newblock in {\em Proceedings of the IEEE conference on computer vision and
  pattern recognition}, 2017, pp. 7263--7271.

\bibitem{redmon2016you}
Joseph Redmon, Santosh Divvala, Ross Girshick, and Ali Farhadi,
\newblock ``You only look once: Unified, real-time object detection,''
\newblock in {\em Proceedings of the IEEE conference on computer vision and
  pattern recognition}, 2016, pp. 779--788.

\bibitem{liu2016ssd}
Wei Liu, Dragomir Anguelov, Dumitru Erhan, Christian Szegedy, Scott Reed,
  Cheng-Yang Fu, and Alexander~C Berg,
\newblock ``Ssd: Single shot multibox detector,''
\newblock in {\em European conference on computer vision}. Springer, 2016, pp.
  21--37.

\bibitem{lin2017feature}
Tsung-Yi Lin, Piotr Doll{\'a}r, Ross Girshick, Kaiming He, Bharath Hariharan,
  and Serge Belongie,
\newblock ``Feature pyramid networks for object detection,''
\newblock in {\em Proceedings of the IEEE conference on computer vision and
  pattern recognition}, 2017, pp. 2117--2125.

\bibitem{fu2017dssd}
Cheng-Yang Fu, Wei Liu, Ananth Ranga, Ambrish Tyagi, and Alexander~C Berg,
\newblock ``Dssd: Deconvolutional single shot detector,''
\newblock {\em arXiv preprint arXiv:1701.06659}, 2017.

\bibitem{tian2019fcos}
Zhi Tian, Chunhua Shen, Hao Chen, and Tong He,
\newblock ``Fcos: Fully convolutional one-stage object detection,''
\newblock in {\em Proceedings of the IEEE International Conference on Computer
  Vision}, 2019, pp. 9627--9636.

\bibitem{duan2019centernet}
Kaiwen Duan, Song Bai, Lingxi Xie, Honggang Qi, Qingming Huang, and Qi~Tian,
\newblock ``Centernet: Keypoint triplets for object detection,''
\newblock in {\em Proceedings of the IEEE International Conference on Computer
  Vision}, 2019, pp. 6569--6578.

\bibitem{everingham2010pascal}
Mark Everingham, Luc Van~Gool, Christopher~KI Williams, John Winn, and Andrew
  Zisserman,
\newblock ``The pascal visual object classes (voc) challenge,''
\newblock {\em International journal of computer vision}, vol. 88, no. 2, pp.
  303--338, 2010.

\bibitem{lin2014microsoft}
Tsung-Yi Lin, Michael Maire, Serge Belongie, James Hays, Pietro Perona, Deva
  Ramanan, Piotr Doll{\'a}r, and C~Lawrence Zitnick,
\newblock ``Microsoft coco: Common objects in context,''
\newblock in {\em European conference on computer vision}. Springer, 2014, pp.
  740--755.

\bibitem{oaod}
Javed Iqbal, Muhammad~Akhtar Munir, Arif Mahmood, Afsheen~Rafaqat Ali, and
  Mohsen Ali,
\newblock ``Orientation aware object detection with application to firearms,''
\newblock {\em arXiv preprint arXiv:1904.10032}, 2019.

\bibitem{cao2018openpose}
Zhe Cao, Gines Hidalgo, Tomas Simon, Shih-En Wei, and Yaser Sheikh,
\newblock ``Openpose: realtime multi-person 2d pose estimation using part
  affinity fields,''
\newblock {\em arXiv preprint arXiv:1812.08008}, 2018.

\bibitem{singh2018analysis}
Bharat Singh and Larry~S Davis,
\newblock ``An analysis of scale invariance in object detection snip,''
\newblock in {\em Proceedings of the IEEE conference on computer vision and
  pattern recognition}, 2018, pp. 3578--3587.

\bibitem{huang2017speed}
Jonathan Huang, Vivek Rathod, Chen Sun, Menglong Zhu, Anoop Korattikara,
  Alireza Fathi, Ian Fischer, Zbigniew Wojna, Yang Song, Sergio Guadarrama,
  et~al.,
\newblock ``Speed/accuracy trade-offs for modern convolutional object
  detectors,''
\newblock in {\em Proceedings of the IEEE conference on computer vision and
  pattern recognition}, 2017, pp. 7310--7311.

\bibitem{olmos2018automatic}
Roberto Olmos, Siham Tabik, and Francisco Herrera,
\newblock ``Automatic handgun detection alarm in videos using deep learning,''
\newblock {\em Neurocomputing}, vol. 275, pp. 66--72, 2018.

\bibitem{akcay2018using}
Samet Akcay, Mikolaj~E Kundegorski, Chris~G Willcocks, and Toby~P Breckon,
\newblock ``Using deep convolutional neural network architectures for object
  classification and detection within x-ray baggage security imagery,''
\newblock {\em IEEE transactions on information forensics and security}, vol.
  13, no. 9, pp. 2203--2215, 2018.

\bibitem{simon2017hand}
Tomas Simon, Hanbyul Joo, Iain Matthews, and Yaser Sheikh,
\newblock ``Hand keypoint detection in single images using multiview
  bootstrapping,''
\newblock in {\em CVPR}, 2017.

\bibitem{wei2016cpm}
Shih-En Wei, Varun Ramakrishna, Takeo Kanade, and Yaser Sheikh,
\newblock ``Convolutional pose machines,''
\newblock in {\em CVPR}, 2016.

\bibitem{wang2019deep}
Tiancai Wang, Rao~Muhammad Anwer, Muhammad~Haris Khan, Fahad~Shahbaz Khan,
  Yanwei Pang, Ling Shao, and Jorma Laaksonen,
\newblock ``Deep contextual attention for human-object interaction detection,''
\newblock in {\em Proceedings of the IEEE International Conference on Computer
  Vision}, 2019, pp. 5694--5702.

\bibitem{gkioxari2018detecting}
Georgia Gkioxari, Ross Girshick, Piotr Doll{\'a}r, and Kaiming He,
\newblock ``Detecting and recognizing human-object interactions,''
\newblock in {\em Proceedings of the IEEE Conference on Computer Vision and
  Pattern Recognition}, 2018, pp. 8359--8367.

\end{thebibliography}

\end{document}